\newcommand{\longversion}[1]{#1}
\newcommand{\shortversion}[1]{}
\documentclass[10pt,letterpaper]{article}

\usepackage[hscale=0.7,scale=0.75]{geometry}
\shortversion{%
\usepackage{cogsci}
\cogscifinalcopy 
}
\usepackage{amssymb,amsthm}
\usepackage[fleqn]{amsmath}
\usepackage{array,multirow}
\usepackage{booktabs}
\usepackage{tabularx,colortbl}
\usepackage{pslatex}
\usepackage{apacite}
\usepackage{float}
\usepackage{xspace}
\usepackage{url}

\usepackage {tikz}
\usepackage{mathtools}
\usepackage{tikz-cd}
\usetikzlibrary{trees,snakes}
\usetikzlibrary{decorations.pathmorphing}
\usetikzlibrary{arrows}
\usetikzlibrary{arrows.meta,calc}
\usetikzlibrary {positioning}
\usetikzlibrary{shadows}
\tikzset{
diagonal fill/.style 2 args={fill=#2, path picture={
\fill[#1, sharp corners] (path picture bounding box.south west) -|
                         (path picture bounding box.north east) -- cycle;}},
reversed diagonal fill/.style 2 args={fill=#2, path picture={
\fill[#1, sharp corners] (path picture bounding box.north west) |- 
                         (path picture bounding box.south east) -- cycle;}}
}

\usetikzlibrary{patterns}
\usetikzlibrary{shapes,arrows,chains,positioning}

\def\newblock{\hskip .11em plus .33em minus .07em}

\usepackage{booktabs}
\usepackage{microtype}

\newcommand{\pneg}{\text{not }}
\newcommand{\Pcond}{\CalP_{\textsf{basic}}}\newcommand\rot[1]{\rotatebox{90}{\footnotesize#1}}

\newcommand{\ifasp}{\leftarrow}
\newcommand{\hyp}{\textsf{hyp}}
\newcommand{\query}{\textsf{question}}

\newcommand{\prinfont}[1]{\textsc{#1}}
\newcommand{\princautious}{\prinfont{cautious}\xspace}
\newcommand{\prinminimal}{\prinfont{minimal}\xspace}

\newcommand{\prinallsuf}{\prinfont{all sufficient}\xspace}
\newcommand{\prinindividual}{\prinfont{individuals}\xspace}
\newcommand{\prinexplain}{\prinfont{explain}\xspace}
\newcommand{\prinfact}{\prinfont{fact}\xspace}
\newcommand{\prinnocontra}{\prinfont{consistency}\xspace}
\newcommand{\necessary}{\textsf{necessary}\xspace}
\newcommand{\sufficient}{\textsf{sufficient}\xspace}
\newcommand{\prinnec}{\prinfont{necessary}\xspace}
\newcommand{\prinsuf}{\prinfont{sufficient}\xspace}
\newcommand{\prinhypo}{\prinfont{hypothesis}\xspace}

\newcommand{\prem}{\textsf{prem}}
\newcommand{\nprem}{\textsf{nprem}}
\newcommand{\concl}{\textsf{concl}}
\newcommand{\nconcl}{\textsf{nconcl}}

\newcommand{\CalP}{{P}}
\newcommand{\CalH}{{H}}
\newcommand{\CalB}{{B}}
\newcommand{\CalM}{{M}}

\newcommand{\essay}{\mathit{essay}}
\newcommand{\library}{\mathit{library}}
\newcommand{\nessay}{{\mathit{not\; essay}}}
\newcommand{\nlibrary}{{\mathit{not\; library}}}

\newcommand{\CalE}{{\cal E}}

\newcommand{\CalO}{{\cal O}}

\DeclareMathOperator{\at}{at}

\newcommand{\eqdef}{\coloneqq}
\newcommand{\SB}{\{}%
\newcommand{\SM}{\mid}%
\newcommand{\SE}{\}}%
\DeclareMathOperator{\AS}{AS}
\newcommand{\plaus}{\mathbb{P}}

\newcommand{\Card}[1]{\ensuremath{| #1 |}}
\newcommand{\lconcl}{\ensuremath{\leftarrow}}

\usepackage[ruled,vlined,linesnumbered]{algorithm2e}

\SetKwInput{KwData}{In}
\SetKwInput{KwResult}{Out}
\SetAlFnt{\small}
\SetAlCapFnt{\small}
\SetAlCapNameFnt{\small}
\SetAlCapHSkip{0pt}
\SetEndCharOfAlgoLine{}
\IncMargin{-\parindent}

\usepackage{nicefrac} 
\newtheorem{examplex}{Example}
\newenvironment{example}
{\pushQED{\qed}\examplex}
  {\popQED\endexamplex}

\newtheorem{definition}{Definition}
\usepackage{array}
\newcolumntype{H}{>{\setbox0=\hbox\bgroup}c<{\egroup}@{}}

\newcommand{\citex}[1]{\citeA{#1}}%

\title{ %
  A Quantitative Symbolic Approach to Individual Human
  Reasoning\thanks{Authors are stated in alphabetical order.}
}

\longversion{%
\author{Emmanuelle Dietz,$^\text{a}$ Johannes K. Fichte$^\text{b}$,  Florim Hamiti$^\text{c}$ \quad\quad\bigskip \\
\small $^a$ Airbus Central Research \& Technology,
Hamburg, Germany, emmanuelle.dietz@airbus.com  \\
\small  $^b$ TU Wien, Vienna, Austria, johannes.fichte@tuwien.ac.at \\
\small  $^c$ florim.hamiti@fit.fraunhofer.de
  }

\date{}
}

\begin{document}

\maketitle

\begin{abstract}
  {Cognitive theories for reasoning} are about understanding how humans come to conclusions from a set of premises. Starting from hypothetical thoughts, we are interested which are the implications behind basic everyday language and how do we reason with them. 
  A widely studied topic is whether cognitive theories can account for 
  typical reasoning tasks and be confirmed by own empirical
  experiments. 
 This paper takes a different view and we do not propose a theory, but instead take findings from the literature and show how these, formalized as cognitive principles within a logical framework, can establish a quantitative notion of reasoning, which we call plausibility.
For this purpose, we employ techniques from non-monotonic reasoning and
  computer science, namely, a solving paradigm called answer set
  programming (ASP).
  Finally, we can fruitfully use plausibility reasoning in ASP to test
  the effects of an existing experiment and explain different
  majority responses.
\smallskip \\

\end{abstract}

\section{Introduction}\label{s:intro}
Usually, the adequacy of cognitive reasoning theories is assessed with respect to typical reasoning tasks,~e.g.,~\cite{byrne:89,wason:68} and own experiments.
The aim is to understand how, from a hypothetical
thought, humans reason and make conclusions. For example, given conditionals such as ``if A, then B''
together with a set of given premises, we can ask what humans conclude
from this information. The adequacy of a cognitive theory is assessed by how well it can account the human data.
Over the decades, many theories have been
proposed~\cite{johnsonlaird:1983,rips:1994,polketal:1995,chater:oaksford:1999,stenning:vanlambalgen:2008,hk:2009a}.
Here, we briefly discuss two dominant theories.
The Probability Heuristics Model (PHM) is a cognitive theory where the environment is described by prior probabilities and updates are done according to Bayes' theorem~\cite{chater:oaksford:1999}.
PHM does not suggest how probabilities are computed, i.e.\ no implemented algorithm exists~\cite{probofcond:2021}.
The (Mental) Model Theory~\cite{johnsonlaird:1983} %
 assumes that humans reason by constructing and manipulating
mental models, which illustrate the possibilities of how the world is perceived by the reasoner~\cite{khemlani:johnsonlaird:2013}.
The model theory with naive probabilities~\cite{johnsonlaird:etal:1999} provides a simple algorithm without using Bayes's theorem, for computing subjective probabilities %
and was further extended~\cite{khemlani:johnson-laird:2015,khemlani:2016}.
These two theories seem to have conflicting viewpoints~\cite{oaksfordetal:2020,knauffetal:2021,oaksford:2021,over:2021} and so far, there is no agreement on whether an integration is possible~\cite{probofcond:2021,over:2021}.

So far, a widely accepted framework for cognitive reasoning does not exist. Even though there might be some agreement on the metrics for a good theory,~e.g., generalizability~\cite{thomsonetal:2015}, simplicity, and predictive accuracy~\cite{taatgenanderson:2010}, %
 theories are not always formalized by their inventors and thus not applicable to tasks straightaway. When effort is done to make them testable and accessible to others,~e.g.,~\cite{khemlani:2012}, the theory might be ambiguously understood and not adequately modeled,~e.g.,~\cite{baratgin:2015}. 
Quite an example of Newell's observation---even though scientists make
excellent research, they \textit{never seem in the experimental
literature to put the results of all the experiments
together}, which obstructs progress~\cite{newell:1973}.
Among others, Newell suggested developing complete processing models, and a computer system that can perform all mental tasks.
On the architectural level, the common model of cognition was proposed, that depicts the \textit{best consensus given the community's understanding of the mind}~\cite{laird:etal:2017}.
  
An additional challenge for cognitive reasoning is \textit{to identify the relevant problems} that a model should account for~\cite{ragni:2020}. Therefore, Ragni~\citeyear{ragni:2020} suggested establishing generally accepted benchmarks, similar to the PRECORE Challenge~\cite{ragni:precore:2019} for human reasoning tasks. 
The evaluation of this challenge was done with the benchmarking tool Cognitive COmputation for Behavioral Reasoning Analysis (CCOBRA) framework~\cite{RiestererShadownox21}.

In other disciplines such as mathematics and computer science, annual problem challenges, such
as the famous DIMACS challenges~\cite{Johnsonothers1990},
SAT~\cite{SatComp21}, and ASP competitions~\cite{gebser:2020}, provided
a community-building tool and contributed to tremendous progress in
actual problem solving. On the side, these efforts result in common
(intermediary) languages. %
At the same time, the outcome of the challenges defines the empirical upper bounds of the state-of-the-art model's performance and determines the performance of new theories~\cite{riesterer:etal:2020}.
Here, we do not propose a cognitive theory but formalize widely accepted findings as task-independent cognitive principles within one framework.
These principles require that assumptions have to
be general enough to be understood in various
contexts. %
At the same time, they call for an unambiguous formalization
that can immediately be instantiated to a specific context.
As
\emph{logical reasoning is $[\ldots]$ considered one of the most  fundamental cognitive activities}~\cite{wolenski:2016},
   a logical formalization of higher-level cognitive assumptions might be suitable, even though not classical logic:
The formalization and reasoning with \emph{default assumptions}, which
are facts that are true in the majority of contexts but not always,
require \emph{non-monotonic
  logic}~\cite{Reiter78,Reiter80a}.
A widely used modeling and problem solving paradigm in AI and computer science that implements non-monotonic
reasoning is \emph{answer set programming
  (ASP)}~\cite{HeuleSchaub2015,GebserKaminskiKaufmannSchaub12}.
The solutions of the program \emph{answer sets} or stable models, can then
be understood as possible models for that program. 

We use the well-established ASP paradigm to model human reasoning
principles, and employ ASP for quantitative reasoning by defining a notion
of plausibility that relates the number of models under assumptions
of interest to the total number of models.
Thereby, we obtain a framework that allows for implementing cognitive
principles. %
In our work, we anti-disciplinary combine findings from cognitive
science, the non-monotonic logic community in computer science, and a
method of model quantification. 
However, it is important to emphasize that ASP is not a theory of cognitive reasoning.

\paragraph{Contributions.}
Our main contributions are as follows:
\begin{enumerate}
\item We show that existing cognitive principles can be well
  represented as rules in ASP following a natural semantics.
  Each program is a set of these rules and yields possible models.
  \item We instantiate these general principles to a well-known task from empirical experiments into simple programs that do plausibility reasoning with ASP.
\item We illustrate, how we can turn potentially multiple models of an ASP program into a quantitative approach to reasoning (plausibility) to test the effects of existing experiments and explain different majority responses.
\end{enumerate}

\section{Preliminaries}
\label{s:asp}

\paragraph{Answer Set Programming (ASP).}
ASP is a popular declarative modeling and problem solving framework in
computer science and artificial intelligence with roots in
non-monotonic
logic~\cite{BrewkaEiterTruszczynski11,GebserKaufmannSchaub12a}.
In ASP, one states problems using propositional atoms, meaning, that an
atom~$a$ can either be true or false. 
A program consists of rules, that state conclusions about
atoms. Solutions to the program are called answer sets (or stable models).
A rule of the form $a \lconcl b, \pneg c$, intuitively, states that
we can conclude $a$ if $b$ is true unless we have evidence that $c$ is
true.
By default, in ASP, we assume that an atom~$a$ is false unless we can
conclude it. This ``in dubio pro reo''-like approach is known as
closed-world assumption (CWA)~\cite{Reiter80a,Reiter78}.
Take the following example.

\begin{example}[False by Default]\label{ex:easy}
  Consider the following conditional sentence.
  \emph{``If it is weekend ($w$), then she will go to the beach
    ($b$)''}.  However, we know that \emph{``She will not go to the
    beach ($b$), if it is cloudy ($c$).''}
  We can rephrase this as follows: \emph{``If it is weekend ($w$),
    then she will go to the beach ($b$) unless it is cloudy ($c$).''}
  This can be modeled as
  program %
  $\CalP_1 = \{b \lconcl w, \pneg c.\}.$ 
 \end{example}
  What are the (intended) models of the programs?
  Here we are interested in the answer sets (or stable models) of the programs, but we will not provide their formal definitions and 
  rather explain the intuition by the next examples. 
\shortversion{%
The interested reader is referred to 
refer to an extended version~\cite{DietzFichteHamiti22b} or introductory
  literature~\cite{GebserKaminskiKaufmannSchaub12}.
}
\longversion{%
The interested reader is referred to 
  Section~\ref{app:defs} of the Appendix. In addition, we suggest to
  consult introductory
  literature~\cite{GebserKaminskiKaufmannSchaub12}.
}
 \begin{example}[Answer Set]\label{ex:answerset}
  The only answer set of~$\CalP_1$ is $\emptyset$, since we neither
  have evidence for cloudy, nor weekend, nor beach.
  If we know that \emph{``it is weekend''}, we take
  program~$P_2= \{b \lconcl w, \pneg c.\;\; w.\}$.
  In $P_2$, we have evidence for weekend by $w$, but no evidence for
  cloudy. From this knowledge, we can conclude beach from the rule
  in~$P_1$. The only answer set of~$\CalP_2$ is $\{w,b\}$.
  In contrast, $P_2= \{b \lconcl w, \pneg c.\;\; w.\;\; c.\}$ has only the answer
  set $\{w,c\}$. We cannot conclude $b$, as we have evidence for $c$
  and the rule~$b \lconcl w, \pneg c$ contains $c$ as an exception to draw the
  conclusion.
\end{example}

  In ASP, we can also make explicit choices to set an atom to true or
  not, which we illustrate in the following example.
\begin{example}[Choices]\label{ex:choice}
Take program~$P_1$ from Example~\ref{ex:easy}. 
  If we know that it could either be weekend or not weekend, we add a
  choice rule to our program.
  \emph{A choice rule states that any combination of atoms inside the
    set are true, including none.}\footnote{Sometimes we also write
    $n\{a_1; a_2;\ldots;a_k\}m$, meaning that we chose at least~$n$
    atoms and at most~$m$ atoms in the choice.}
  We obtain
  $P_3 = \{b \lconcl w, \pneg c.\;\;\; \{w\}. \}$.
  Then, $P_3$ has two answer sets~$\{w,b\}$ and $\{\}$.
\end{example}

Next, we illustrate how adding rules can effect conclusions.
%
%
%
%
%
%
%
%
%
%
\begin{example}
\label{ex:running}
Consider program\\
%
$\CalP_4 = \{ %
  \underbrace{b \lconcl w, \pneg c.}_{r_1}\; %
  \underbrace{\{w\}.}_{r_2}\; %
  \underbrace{\{s; c\}.}_{r_3}\; %
  \underbrace{c \lconcl \pneg s}_{r_4} %
  \}$
\\
  where $r_1$ corresponds to the conditional in Ex.~\ref{ex:easy}, $r_2$ to the choice in
  Ex.~\ref{ex:choice}, and $s$ means \emph{``it is sunny''}.  Choice
  $r_3$ states that either $s$ or $c$ are true, both are
  true, or none is true.
%
We have six answer sets,
namely,~$\AS(\CalP_4) = \SB \{s\}, \{c\}, \{c,s\}, \{b,s,w\}, \{c,s,w\},\{c,w\}\SE$.  The
set~$X=\{b,w\}$ is not an answer set of~$P_4$, since we conclude cloudy
($c$) from rule~$r_4$ if we have no evidence for sunny ($s$).
\end{example}%
Below, we also use variables in the programs, which provides us with a
more expressive and compact way of representation. We omit formal
details for space reasons, but again give an example. By
$\{p(X).\, r(X) \lconcl p(X)\}$ for $X \in \{a,b\}$, where $a$ and $b$
are constants, we mean
$\{p(a).\, p(b).\, r(a) \lconcl p(a).\, r(b) \lconcl p(b)\,\}$.
We assume that there is always at least one constant
in~$\CalP$ and $\AS(\CalP)$ is  the set of all answer sets
of~$P$.

\paragraph{Quantitative Reasoning in ASP.}
Traditionally, when modeling in logic, one considers simple decision
questions,~i.e., yes-no questions~\cite{CopiCohenRodych2019}. In terms of ASP
this would simply mean asking whether a given program has an answer
set,~i.e., $\AS(\CalP) \neq \emptyset$.
Beyond, we find questions such as credulous and skeptical reasoning. Where
\emph{credulous} or \emph{skeptical reasoning} asks whether an atom~$a$ is
contained in \emph{at least one} and \emph{all} answer sets,
respectively.
We are also interested in computing the \emph{plausibility} for a
set~$Q$ of rules %
relating to the number of answer sets under the assumption of the total
number of answer sets.

\begin{definition}[Plausibility]\label{def:projplaus}
  Let~$P$ be a program and $Q$ be a set atoms, called
  \emph{questions}. Then, \emph{plausibility} of~$\CalP$ under~$Q$ is
  defined~$\plaus[\CalP, Q] \eqdef\frac{\Card{\AS(\CalP \cup
      \CalP_Q)}}{\max(1,\Card{\AS(\CalP)})}$ where $\CalP_Q$ consists
  of integrity rules that ask whether atoms in~$Q$ can be made
  true,~i.e., $\CalP_Q \eqdef \{ \lconcl \pneg a \SM a \in Q\}$.
\end{definition}
Later, when representing questions in ASP programs, we assume that~$Q$
is given by rules of the form~``$\text{question}(a).$'' for
every~$a \in Q$. When computing $\Card{\AS(P \cup Q)}$, we replace
each $\text{question}(a)$ as above by ``$\lconcl \pneg a.$''.
By using ASP, we describe the system and the outcome using rules
within ASP. Answer sets represent the outcomes of our system. The question
for plausibility still relates to inference and is done in terms of
counting answer sets.
Using modern implementations that solve answer set programs, we can
obtain the plausibility by listing all answer sets and computing the
relation.
The problem is of high computational complexity and has only recently
received more attention with the rise of more efficient solving
techniques in the propositional setting that avoid
enumeration~\cite{LagniezMarquis17a,SharmaEtAl19a,FichteHecherHamiti20,FichteEtAl20,FichteHecherRoland21}
or in the ASP
setting~\cite{KabirEtAl22,FichteGagglRusovac22,NadeemFichteHecher22}.

\begin{example}
  Consider~$\CalP_4$ from Ex.~\ref{ex:running}. 
  For skeptical reasoning for $s$ or $c$ in $\CalP_4$, the outcome is no. Whereas for credulous reasoning, the outcome is yes. When considering the plausibility of $P_4$
  under $s$ being true, meaning $\CalP_Q = \{ \lconcl \pneg s\}$, we can
  see that
  $\Card{\AS(\CalP_4 \cup \CalP_Q)} = \Card{\SB \{s\}, \{c,s\}, \{b,s,w\}, \{c,s,w\} \SE} = 4$ and
  $\Card{\AS(\CalP_4)}= 6$,
  which yields $\plaus[\CalP_4,Q] = \nicefrac{4}{6}$.
\end{example}

Below, we illustrate counting and Bayesian views as well as
differences to our notion.  We follow a popular example by
McElreath~\citeyear{McElreath2020,McElreath20201}.
Recall that Bayes-Price theorem is used to compute the probability of
an event, based on prior knowledge of conditions that might be related
to the event. While it might seem quite plain, one can just list
potential combinations and count possible ways instead.

\begin{example}%
\label{ex:bayes}
  Assume that we have a bag of four marbles, which could be blue~(b)
  or white~(w). We are not aware of how many of each is in the bag.
  From the four marbles, the cases %
  (i)~wwww; %
  (ii)~bwww; %
  (iii)~bbww; %
  (iv)~bbbw; and %
  (v)~bbbb %
  are possible.
  To obtain more detailed information about the content, we can take
  one marble remember its color and put it back.
  Assume that after repeating times, we observe bwb.
  To estimate Bayesian plausibility, we can count how many ways are to
  produce each of the Cases~(i-v) assuming the seen data. In more
  detail, 0 ways for wwww, 3 ways for bwww, 8 ways for bbww, 9 ways
  for bbbw, and 0 ways for bbbb. In total 20 possible ways. 
  Plausibility talks about an observation in relation to all possible
  ways. Here, $\nicefrac{0}{20} = 0$ for wwww, $\nicefrac{3}{20}=0.15$
  for bwww, $\nicefrac{8}{20} = 0.40$ for bbww,
  $\nicefrac{9}{20} = 0.45$ for bbbw, and $\nicefrac{0}{20} = 0$ for
  bbbb.
  Our framework allows to express this in our notion of
  plausibility. Therefore, we can model the 5 cases that can be
  produced and their resulting ways of producing the data. Then, ask
  for the number of solutions that can be produced in total and the
  one under the assumption say Case~(ii) bwww.
  \shortversion{%
    We provide a detailed ASP program in an extended
    version~\cite{DietzFichteHamiti22b}. %
  }%
  \longversion{%
    We refer to Listing~\ref{lst:bayes} in the Appendix.
  }%
  While our framework allows to express such questions we are more
  general and by \emph{plausibility in ASP}, we express the relation
  of count under assumption and total count of possible answer sets.
\end{example}

The existing probabilistic approaches to human reasoning differ from our proposal as those probabilities are either understood as subjective and are not derived from the quantification over models,~e.g.~\cite{chater:oaksford:1999} or attach the probabilities to different types of inferences,~e.g.~\cite{Kleiter18}.
We also use a slightly different approach than~\citex{Johnson-Laird99}
by considering the relationship
on counting the values in the truth table that evaluate to true, but
according to the answer set semantics.

\section{Cognitive Principles in ASP}

We employ accepted findings from the
literature and formalize them as rules, called
\emph{cognitive principles}, within one framework and explain their
effects. As a baseline, we consider the principles presented in the literature~\cite{dietz:kakas:2020,dietz:kakas:2021}.
These principles are task-independent and can be any assumption that humans seem to make
regardless of whether they are valid in classical logic.
ASP will be the framework in which we formalize them.
Before we proceed with the rules and their representation in ASP, let us clarify that we do not present a new cognitive theory.

\paragraph{Presuppositions}

\citeauthor{grice:1975}'s~\citeyear{grice:1975} conversational implicatures are about additional interpretations of the sentences we hear, not necessarily related to the content. For instance, we usually communicate according to the cooperation principle. Thus, when the experimenter (or someone we trust) states ``$a$ is true or $a$ is false'',  we assume that this is true. In ASP, a fact $\prem(a)$ is represented as 
either $\prem(a)$ or $\nprem(a)$, respectively (\prinfact principle).\footnote{
Throughout the paper, the negation of a statement $a(X)$ is represented with an auxiliary statement $na(X)$,~i.e., having the same name as the statement, preceded by an 'n'.}
Yet, both cannot be true at the same time (\prinnocontra principle). %

Grice's maxim of relevance implies that everything that is said, seems to be relevant, suggesting that
humans might generate hypotheses from the context (\prinhypo principle).
We account for this principle by establishing context-dependent hypotheses for each statement $a$ that we are made aware of by adding $\hyp(a)$.
 \begin{align}
\{\prem(X);\nprem(X)\}1 \ifasp \hyp(X). 
\tag{\prinhypo}
 \end{align}
The $1$ denotes that at most one statement can be true ensuring the \prinnocontra principle.

\paragraph{Types of Conditionals}
Conditions in conditionals can be of different types, such as necessary or sufficient~\cite{byrne:etal:1999,byrne:2005}.
Consider the two conditional sentences \textit{If she meets with a friend, then she will go to the play} and \textit{If she has enough money, then she will go to the play}.
We assume that \textit{she meets with a friend} is a sufficient condition whereas \textit{she has enough money} is not sufficient but a necessary condition for \textit{she will go to the play}.
Assume that \textit{she meets a friend}. Together with the above \prinhypo principle and given the second conditional, humans might generate the hypothesis that 
 \textit{she does not have enough money} which functions as  a \textbf{disabling condition}~\cite{cummins:1991} to the modus ponens conclusion that \textit{she will go to a play}.
The follwing rule states that $\concl$ follows if condition is asserted to be true (modus ponens):
 \begin{align}
\concl \ifasp \prem(X), \sufficient(X). & \tag{\prinsuf}
\end{align}
The following rule states that $\nconcl$ follows if condition is false (denial of the consequent).
 \begin{align}
\nconcl \ifasp \nprem(X), \necessary(X).& \tag{\prinnec}
\end{align}

Let us observe that \textit{she does not meet a friend}. If this condition is also necessary, according to the \prinnec principle we might conclude that \textit{she will not go to the play}.
Consider now additionally that \textit{if she has free tickets, she will go to the play}. The hypothesis that \textit{she has free tickets} functions as an \textbf{alternative cause}~\cite{cummins:1991} to the condition \textit{she meets a friend}, for the conclusion \textit{she will go to the play}.
The following rule captures this idea:
 \begin{align}
\nconcl \ifasp \nprem(X_1), \dots , \nprem(X_n) & \tag{\prinallsuf}
\end{align}
where $\nprem(X_1), \dots , \nprem(X_n)$ is the conjunction of all $X_i$, $1 \leq i \leq n$,
for which there exists a rule of the following form: $\concl \ifasp \prem(X_i), \sufficient(X_i)$.
This rule states that $\nconcl$ follows when \textit{all} its sufficient conditions are false.

\paragraph{Maxim of Inference to the best explanation}

Even though not valid in classical logic, humans have the ability to reason from observations to explanations, called abduction~\cite{peirce:1903}. As reported by~\citex{kelley:1973,Sloman94}, contrastive (or alternative) explanations might increase or decrease their plausibility, depending on the context.

 In ASP, abduction can be implemented as cautious (or skeptical) abduction~\cite{kakas:etal:1993}. Given a program $\CalP$ and an observation $\CalO$, is $\CalE$ an explanation for $\CalO$? This question can be answered in a two-step procedure: %
(i) Generate models of $\CalP$, in which $\CalO$ holds: $\ifasp \CalO. \hfill (\prinexplain \text{ principle})$ %
(ii) Select models in which $\CalE$ holds: $\query(\CalE). (\princautious  \text{ principle})$\\
We additionally require explanations to be minimal (\prinminimal principle): Given $\CalP$, $\CalE$ is a minimal explanation of $\CalO$ if and only if there is no other explanation $\CalE'$ for $\CalO$ such that $\CalE' \subset \CalE$.
In the sequel,
$\CalO$ is either $\concl$ or $\nconcl$ and $\CalE$ is either $\prem$ or $\nprem$. Given that \textit{if $\prem$ then $\concl$}, the derivation from $\concl$ to $\prem$ corresponds to the (classical logically) invalid affirmation of the consequent,
whereas the derivation from $\nconcl$ to $\nprem$ corresponds to the valid modus tollens.

\paragraph{Individual Reasoners}

Humans differ in their reasoning,~c.f.,~\cite{khemlani:2016}.
We represent these differences as choice rules, surrounded by $\{ \dots \}$, which can contain one or more variables. For instance,
models in which $\hyp(a)$ is true, false, or unknown, can be generated through the choice rule ``$\{ hyp(a)\}$''  (\prinindividual principle).

\section{Application to Human Reasoning}

We discuss the application of cognitive principles within ASP by means
of a typical reasoning task. The suppression task~\cite{byrne:89}
consists of two parts, where participants were divided into three groups and were asked whether they
could derive conclusions given variations of a set of premises.
First, we present the formalization in ASP guided by cognitive
principles. In Part~I, reasoning is done deductively, and, in Part~II,
it is done abductively. In contrast to other logic programming approaches~\cite{stenning:vanlambalgen:2008,dhr:2012}, we apply
quantitative reasoning to the computed models which allows us to account for the majority's differences in the experimental results.

\begin{table}
\centering
\begin{tabular}{@{\hspace{0cm}}l@{\hspace{0.2cm}}l@{\hspace{0.1cm}}l@{\hspace{0.1cm}}l@{\hspace{0.1cm}}l}
\toprule
Case & All Groups & Group I & Group II & Group III \\ 
\midrule
All cases & $\Pcond$ & $\{\necessary(e)\}$ &  & $\necessary(e)$\\\midrule
$\essay$       & $\query(\concl)$ & $\prem(e)$ & $\prem(e),$ & $\{\hyp(o);\prem(e)\}$\\
                         &                                    &                       & $ \{\hyp(t)\}$ \\
\midrule
$\mathit{not}$      & $\query(\nconcl)$, &  & $\{\hyp(t)\}$ & $\{\hyp(o)\}$\\
  $\essay$                         &  $\nprem(e)$ \\
\midrule
 $\library$    & $\{\concl\}, \ifasp \pneg \concl,$ & $\hyp(e)$ & 
$\{ \hyp(e);\hyp(t)\}1$ &$\hyp(o), \hyp(e)$\\
                       & $ \query(\prem(e))$ \\
\midrule
$\mathit{not}$      & $\{\nconcl\}, \ifasp \concl,$  & $\hyp(e)$ & 
$\hyp(e), \hyp(t)$ & $\{\hyp(o);\hyp(e)\}1$ \\
 $\library$                       & $\query(\nprem(e))$ \\
\bottomrule
\end{tabular}
\caption{\label{tab:programs}
Summary of the rules applied by case and group. The second row shows the rules by group that applied to all cases.
}
\end{table}

\paragraph{Part I: Search for conclusions}

Group I was given the following two premises:
\textit{If she has an essay to finish, then she will study late in the library. She has an essay to finish.}  ($\essay$)
The participants were asked what of the following answer possibilities follows assuming that the 
above premises were true:
\textit{She will study late in the library, She will not study late in the library.}  or
\textit{She may or may not study late in the library.}
96\% of the participants in this group concluded that \textit{She will study late in the library} ($\library$).
 Group II of participants additionally received the following premise:
\textit{If she has a textbook to finish, then she will study late in the library}, 
 which yields to the same result: 
96\%\footnote{We refer to the percentages from Byrne~\citeyear{byrne:89}. Table~\ref{tab:results} also shows the percentages from Dieussaert~\citeyear{dieussaert:2000}.} of the participants in this group concluded that \textit{She will study late in the library}.
Group III of participants instead additionally received the following premise:
\textit{If the library is open then she will study late in the library.}
In this case, only 38\% concluded that \textit{She will study late in the library}.
Even though the conclusion was logically valid in all three groups (modus ponens), a \textit{suppression effect} in Group III could be observed~\cite{byrne:89}. This effect very well demonstrates the non-monotonic nature of human reasoning.

If instead, \textit{She does not have an essay to finish} was given as a fact, only 4\% of Group II concluded \textit{She will not study late in the library}, whereas for Group I and Group III, it was 46\% and 63\%, respectively. Here, the conclusion was not valid (affirmation of the consequent), and the suppression effect could be observed in Group II. This case nicely shows that the suppression effect occurs independent on whether the conclusion is valid.

Motivated by the cognitive principles, the following rules, denoted by $\Pcond$, are part of all cases and groups:
\begin{align}
&\ifasp \concl,\nconcl.		  \tag{\prinnocontra} \\
&\concl \ifasp \prem(X), \sufficient(X).	 & \tag{\prinsuf} \\
&\nconcl \ifasp \nprem(X), \necessary(X). & \tag{\prinnec} \\
&\{\prem(X); \nprem(X)\}1 \ifasp \hyp(X).		& \tag{\prinhypo} \\
&\nconcl \ifasp \nprem(e), \dots , \nprem(t) & \tag{\prinallsuf} \\
&\sufficient(e). & \tag{\prinsuf} \\
&\sufficient(t). & \tag{\prinsuf} \\
&\necessary(o) . & \tag{\prinnec}
\end{align}
$\concl$ and $\nconcl$ here refer to \textit{She will study late in the library} and \textit{She will not study late in the library}, respectively.
The last three rules state that $e$
(\textit{She has an essay to finish}) and $t$ (\textit{She has a textbook to read}) are sufficient for $\concl$, whereas $o$ 
(\textit{The library is open}) is necessary for $\concl$.
Table~\ref{tab:programs} shows all the programs for all the cases and groups.

Consider the first three rows in Table~\ref{tab:programs}. For different groups of participants different underlying principles are assumed:
By the \prinfact principle for case $\essay$ 
and $\nessay$ we assume $\prem(e)$ and $\nprem(e)$, respectively. 
Similar to the answer possibilities that were given to the participants, in ASP we ask the 
program whether an answer follows by
 $\query(\concl)$ or $\query(\nconcl)$.
The different groups are made aware of different contexts, which is represented by the \prinhypo principle:
 The program for Group II can build the hypothesis $\hyp(t)$, whereas Group III can build the hypothesis $\hyp(o)$.
To account for different participants ($\prinindividual$ principle), choice rules (rules surrounded by $\{ \dots \}$) are used:
Consider $\{\necessary(e)\}$ in Group I for all cases: It allows the generation of models in which $\necessary(e)$ is true, false, or unknown. 
Choice rules enable us to deal with conditions that might result in conflicting conclusions. 
Consider $\{\hyp(o), \prem(e)\}$ (in case $\essay$, Group III). Assume $\hyp(o)$: Because $\necessary(o) \in \Pcond$, by the \prinnec principle $\nconcl$ follows. If we assume $\prem(e)$, as $\sufficient(e) \in \Pcond$, by the \prinsuf principle, $\concl$ follows. 

\paragraph{Part II: Search for Explanations} 

The second part of the experiment was similar, except that the given facts were different.
In the first case, participants were asked what follows, given the fact that
\textit{She will study late in the library} ($\library$). 
For Group I and III, 71\% and 54\% derived the non-valid (affirmation of the consequent) conclusion that \textit{She has an essay to finish}, whereas the suppression effect occurred for Group II, with only 13\%.
In the second case, they were asked what follows, given the fact that
\textit{She will not study late in the library} ($\nlibrary$).
Here, 92\% and 96\% of participants in Group I and II derived the (logically valid) modus tollens conclusion \textit{She does not have an essay to finish}, whereas the suppression effect occurred for Group III, with only 33\%.

Following the \prinexplain principle, participants might have understood the given fact as an observation and searched for explanations.
 $ \ifasp \pneg \concl$  generates all models in which $\concl$  holds and $\query(\prem(e))$ selects the models in which $\prem(e)$ holds; similar for $\nconcl$. These models are explanations for the given observation.

To account for different participants, we specify choice rules:
 $\{\concl\}$ allowing to generate models in which $\concl$ is either false, true, or unknown. The cases in which $\concl$  is simply assumed to hold, represents participants who possibly did not search for explanations or generated other explanations based on their background knowledge. Similar for $\nconcl$.

Consider the special cases of choice rules for the generation of explanations in row 5 and 6 in Table~\ref{tab:programs}:
For Group II, case $\library$,
$\{ \hyp(e);\hyp(t)\}1$ excludes the cases where both 
$\hyp(e)$ and $\hyp(t)$ are true. As both $e$ and $t$ are \textit{sufficient conditions} for $\library$, it is enough to assume either $\prem(e)$ or $\prem(t)$ to hold as an explanation for $\library$.
For Group III, case $\nlibrary$,
$\{ \hyp(e);\hyp(o)\}1$ excludes the cases where both 
$\hyp(e)$ and $\hyp(o)$ are true. As both $e$ and $o$ are \textit{necessary conditions} for $\library$, it is enough to assume either $\nprem(e)$ or $\nprem(o)$ to hold as explanation for $\nlibrary$.

Note that for both groups this rule is not relevant for the other cases. In Group II, both $\nprem(e)$ and $\nprem(t)$ need to hold to be an explanation for $\nlibrary$ whereas in Group III, both $\prem(e)$ and $\prem(o)$ need to hold to be an explanation for $\library$.
These choice rules motivated by the \prinminimal principle are case-specific. Computing minimal explanations is expensive in general~\cite{EiterGottlob1993}.

\paragraph{From Counting Models to Plausibility}
 
Table~\ref{tab:results} shows the number of generated models according to the given programs in Table~\ref{tab:programs} including the plausibility for $\library$, $\nlibrary$, $\essay$, and $\nessay$, respectively.\footnote{The programs, models and the results can be found online: \url{https://github.com/eadietz/bst2asp}}

The plausibility is computed via ASP as described in the preliminaries. 
For each group (column 2) the number of all models for the program (column 4) and the number of all models that satisfied the question (column 3) are depicted. 
Columns 5 to 7 show the computed plausibility of quantitative ASP, the experimental results in the literature~\cite{byrne:89} and~\cite{dieussaert:2000}, respectively.
ASP does not only model well the suppression effect in all four cases but also accommodates for the difference between high percentages (Group I and II for the cases $\essay$ and $\nlibrary$) and significant percentages (Group I and III for case $\nessay$ and $\library$).
Interestingly, whenever a suppression effect occurs in a group, ASP also generates more models compared to the other groups.
This seems to agree with the assumption that inferences which leads to multiple models should be more difficult than the ones on a single model~\cite{probofcond:2021}.

\begin{table}[!ht]
\centering
    \begin{tabular}{H@{\hspace{0cm}}lHlllrr@{\hspace{0.2cm}}r}
\toprule   & & & & \multicolumn{2}{c}{models} &  
\\
& Cases &  & Group & Question & total & ASP & \citeA{byrne:89} & \citeA{dieussaert:2000} \\
\midrule
6  & &  ex1\_1.lp &     I &               2 &             2 &       100 &        96 &        88     \\
7  &  &  ex1\_2.lp &    II &               4 &             4 &       100 &        96 &        93    \\
 \rowcolor{black!20} & \cellcolor{white}  \multirow{-3}{*}{\rot{$\essay$}} & & III &  {3} &             {7} &        43 &        38 &        60 
\\ \cmidrule{3-9}
  & \multicolumn{8}{l}{$\rightsquigarrow$ concluded \textit{She will study late in the library}}
  \smallskip\\ \midrule
9  &  &  ex2\_1.lp &     I &               1 &             2 &      50 &         46 &         49          \\
\rowcolor{black!20} 10 &\cellcolor{white}   &  ex2\_2.lp &    II &               1 &             4 &     25 &          4 &         22  \\
11 & \multirow{-3}{*}{\rot{$\nessay$}}   &  ex2\_3.lp &   III &               5 &             8 &      63 &         63 &         49  
\\ \cmidrule{3-9}
  & \multicolumn{8}{l}{$\rightsquigarrow$ concluded \textit{She will not study late in the library}}
  \smallskip\\ \midrule
0  &  &  ex3\_1.lp &     I &               2 &             5 &     40 &      71 &      53 \\
\rowcolor{black!20}1  & \cellcolor{white}   &  ex3\_2.lp &    II &               1 &             7 &     14 &      13 &      16  \\[0.2em]
2  & \multirow{-3}{*}{\rot{\small$\library$}}     &  ex3\_3.lp &   III &               2 &             4 &     50 &      54 &      55  
\\ \cmidrule{3-9}
  & \multicolumn{8}{l}{$\rightsquigarrow$ concluded \textit{She has an essay to finish}}
  \smallskip\\ \midrule
3  &  &  ex4\_1.lp &     I &               1 &             1 &    100 &       92 &       69 \\
4  &&  ex4\_2.lp &    II &               1 &             1 &     100 &       96 &       69 \\
\rowcolor{black!20}5  &\cellcolor{white}  \multirow{-3}{*}{\rot{\small$\nlibrary$}} &  ex4\_3.lp &III &1&2& 50 &       33 &       44 
\\ \cmidrule{3-9}
  & \multicolumn{8}{l}{$\rightsquigarrow$ concluded \textit{She does not have an essay to finish}}\\
\bottomrule
\end{tabular}
\caption{\label{tab:results}
The results in ASP compared to the experimental results
  The first two columns refer to cases and groups.
  Columns 3 and 4 refer to the number of models that satisfy the question and all models of the program.
  }

\end{table}
\subsection{Discussion and Outlook}
To the best of our knowledge, no approach for human reasoning has considered quantitative model counting.
Additionally, we provide an online accessible formalization of the task such that the results can be replicated.
We believe that a good model needs to account for the assumptions of various theories, for individual reasoners, and can rigorously be applied to benchmarks. An approach, which %
is guided by general cognitive principles, can account for individuals, is rigorously applicable as shown in other domains, and, as motivated by established theories, likely accounts for a variety of tasks.

The computed plausibility in this paper is solely based on the number of models, ignoring the quality of the respective models.
However, some models might be easier to be considered by humans than other models~\cite{knauffetal:1998,ragnietal:2006}, meaning that they are not equiprobable.
An additional preference relation, either on the rule level or on the model level, can easily be implemented in ASP~\cite{brewkaD0S15}, could account for these differences or weighted counting~\cite{SangBeameKautz05a}.

\section{Conclusion and Future Work}
In this work, we showed how model human reasoning
principles can be formalized within answer set programming (ASP), which is a popular
modeling problem, and reasoning framework in artificial intelligence
(AI).
By counting answer sets, we establish a notion of quantitative
reasoning in terms of plausibility and account for 
different majority responses in cognitive reasoning.
While the constructed models were guided by cognitive principles, we
clearly do not believe that human reasoning works similarly as ASP
computation. Instead, ASP helps to represent principles. 

Putting our results into the light of Newell's considerations on
progress within the cognitive community~\cite{newell:1973}, our work
might be seen as yet another framework for mental modeling. However,
we use well-established techniques from AI for representing cognitive
principles and making small steps to converge. %
Thereby, we incorporate existing approaches and open ASP to
the cognitive theory community.

In addition, we aim to investigate whether preferences over answer
sets or weighted counting
could allow for more detailed modeling of cognitive
principles. 
Furthermore, inspired by our idea of employing existing techniques from
AI, and as already mentioned in the introduction,
the cognitive science community could
discuss and design an event establishing benchmarks for human reasoning tasks as suggested in~\cite{ragni:2020}
explaining different majority responses using one or many existing
frameworks.

\section{Acknowledgements}
Work has partially been carried out while the second author was
visiting the Simons Institute at UC Berkeley. The authors gratefully
acknowledge the valuable feedback by the anonymous reviewers.

\bibliographystyle{apacite}

\setlength{\bibleftmargin}{.125in}
\setlength{\bibindent}{-\bibleftmargin}

\longversion{%
\cleardoublepage
\appendix
\section{Omitted Definitions}

\subsection{Answer Sets}
 \label{app:defs}
We follow formal definitions on propositional
ASP~\cite{BrewkaEiterTruszczynski11,JanhunenNiemela16a}.
Let $m$, and $n$ be non-negative integers.
A \emph{program}~$\CalP$ is a finite set of \emph{rules} of the form
\( a \lconcl b_{1}, \ldots, b_{m}, \pneg c_{1}, \ldots, \pneg c_n,
\) 
called \emph{normal}, where $a$ is an atom or $\bot$ and $b_1$,
$\ldots$, $b_m$, $c_1$, $\ldots$, $c_n$ are distinct atoms. For a
normal rule~$r$, if $a=\bot$, then we call $r$ an \emph{integrity
  constraint} and omit $\bot$.
For a rule~$r$, we let $\CalH_r \eqdef \{a\}$,
$\CalB^+_r \eqdef \{a_{1}, \ldots, b_{m}\}$, and
$\CalB^-_r \eqdef \{c_{1}, \ldots, c_n\}$.
In addition, we say a \emph{choice rule}\footnote{A choice
  $1\{a_1; a_2;\ldots;a_n\}$ is simply a shorthand notation for adding
  another integrity constraint of the
  form~$\lconcl \pneg a_1, \ldots \pneg a_n$. By $\{a_1; a_2\}1$, we
  mean that at most one atom can be chosen, which is a short for
  integrity constraints of the form~$\lconcl a_1, a_2$.}
is of the form~$\{a_1; \ldots; a_\ell\}$ where $a_1$, $\ldots$,
$a_\ell$ are distinct atoms, $\CalH_r\eqdef\{a_1, \ldots, a_\ell\}$,
$\CalB^+_r\eqdef \emptyset$, and $\CalB^-_r\eqdef \emptyset$.
If $\CalB^+_r=\CalB^-_r = \emptyset$, we say that $r$ is a \emph{fact} and
omit the symbol~$\lconcl$.
We denote the sets of \emph{atoms} occurring in a rule~$r$ or in a
program~$\CalP$ by $\at(r) \eqdef \CalH_r \cup \CalB^+_r \cup B^-_r$ and
$\at(\CalP)\eqdef \cup_{r\in \CalP} \at(r)$.
A set $\CalM$ of atoms \emph{satisfies} a rule $r$ if
$(\CalH_r \cup \CalB^-_r) \cap \CalM \neq \emptyset$ or
$\CalB^+_r \setminus \CalM \neq \emptyset$. $M$ is a \emph{model} of $\CalP$ if it
satisfies all rules of~$\CalP$.
The \emph{Gelfond-Lifschitz (GL) reduct} of~$P$ under~$M$ is the
program~$P^M \eqdef \SB H_r \lconcl B^+_r \SM r \in P, \text{ r is
  normal}, B^-_r\cap M \neq \emptyset \SE \cup \{a \lconcl \SM r \in
\CalP, \text{ r choice rule}, a \in H_r \cap M
\}$. %
$M$ is an \emph{answer set} of a program~$\CalP$ if (i)~$M$ is a model
of~$\CalP$ and (ii)~$M$ is a minimal model of~$\CalP^M$. Observe that $\CalP^M$
ignores choice rules and by \emph{minimal} we mean that there exists
no~$N \subsetneq M$.

\begin{example}%
\label{ex:longrunning}
Consider program
\[
\CalP = \{ %
  \overbrace{b \lconcl w, \pneg c.}^{r_1}\; %
  \overbrace{\{s; c\}.}^{r_2}\; %
  \overbrace{c \lconcl \pneg s}^{r_3} %
  \}
\]
We have three answer sets,
namely,~$\AS(\CalP) = \SB \{s\}, \{c\}, \{s, c\}\SE$.
The set~$M_1 = \{s\}$ satisfies $r_1$ as
$B^-_{r_1}\setminus M = \emptyset$, $r_2$ as
$H_{r_2} \cap M \neq \emptyset$, and $r_3$ as
$B^-_{r_3} \cap M \neq \emptyset$.
We have the GL-reduct $\CalP^{M} = \SB r_1.\; r_2.\; s \lconcl \SE$, for
which $M_2= \{c\}$ is a minimal model.
Observe that for $M_3 = \{s,c\}$, we obtain the GL-reduct
$\CalP^{M_3}= \SB r_1.\, r_2.\, s.\, c \SE$ for which $M_3$ is a minimal
model.
However, if we consider the
program~$\CalP' \eqdef P \cup \SB \lconcl \pneg c\SE$, the
set~$M_2$ does not satisfy the newly added rule and $M_2$ is
not an answer set of~$\CalP'$.
\end{example}%

\subsection{Encoding Example~\ref{ex:bayes}}

Consider Example~\ref{ex:bayes} from page~\pageref{ex:bayes}. In order to express the
notion of plausibility in ASP, we can develop the program~$P_b$ given
in Listing~\ref{lst:bayes}.
For convenience and to allow for direct executability, we follow
syntax as used in the clingo system~\cite{Potassco22} instead of
mathematical notion.
The program expresses the basic knowledge about the setting in
Lines~\ref{line:step} and \ref{line:data}, meaning, we have three
steps, four marbles, two colors, and we have the data about each step
and color.
Line~\ref{line:cases} expresses the possible cases and the following
Lines~\ref{line:wwww} to \ref{line:bbbb}, which cases are implied by
each selection, meaning which combination we conjecture. For example,
select(1) implies that we conjectured wwww expressed by the
conjecture(1,w), conjecture(2,w), conjecture(3,w), conjecture(4,w).
Line~\ref{line:consistency} ensures that data, conjectured marbles,
and steps are consistent, meaning that for each data(S,C) at step~S
and color~C, we chose exactly one step(S,M,C), so step S with marble M
and color C, where marbles may be obtained from the conjecture.
Then, the answer sets of Program~$P_b$ express the possible marbles in
the bag that are consistent with the seen data.
We can obtain the count by running \texttt{clingo -n0} on the given
program.
If we are in addition interested in the plausibility for one
combination, say bwww, we can simply add the rule~``:- not
select(2).'' and run again.
Counting once under assumption and once without the assumption yields
$\nicefrac{3}{20}$ agreeing with the Bayesian plausibility in this
case.

\begin{algorithm}[h]
\ttfamily
step(1..3). marble(1..4). color(b). color(w).\;\label{line:step}

\% seen data\;
data(1,b). data(2,w). data(3,b).\label{line:data}\;\;

\% Possible Cases\;
1$\{$select(1..5)$\}$1. \label{line:cases}\;\;

\% wwww\;
conjecture(1..4,w) :- select(1).\label{line:wwww}\;
\% bwww\;
conjecture(1,b) :- select(2).\;
conjecture(2..4,w) :- select(2).\;
\% bbww\;
conjecture(1..2,b) :- select(3).\;
conjecture(3..4,w) :- select(3).\;
\% bbbw\;
conjecture(1..3,b) :- select(4).\;
conjecture(4,w) :- select(4).\;
\% bbbb\;
conjecture(1..4,b) :- select(5).\label{line:bbbb}\;\;

\% Ensure consistency with data\;
1$\{$step(S,M,C):conjecture(M,C)$\}$1 :- data(S,C).\label{line:consistency}\;
\rmfamily
\caption{An ASP program~$P_{b}$ that expresses the possible models
  that we can obtain from Example~\ref{ex:bayes}.}
\label{lst:bayes}
\end{algorithm}

}

\end{document}

